\pdfoutput=1
\documentclass[11pt]{article}

\usepackage[preprint]{acl}

\usepackage{times}
\usepackage{latexsym}

\usepackage[T1]{fontenc}
\usepackage[utf8]{inputenc}

\usepackage{microtype}

\usepackage{inconsolata}

\usepackage{graphicx}

\usepackage{todonotes}

\usepackage{times}
\usepackage{latexsym}
\usepackage{amsmath}
\usepackage{graphicx}
\usepackage{array}
\usepackage{booktabs}
\usepackage{tabularx}
\usepackage{tabu}
\usepackage{adjustbox}
\usepackage{enumitem}
\usepackage{longtable}
\usepackage{makecell}

\usepackage[T1]{fontenc}
\usepackage[utf8]{inputenc}

\usepackage{microtype}

\usepackage{inconsolata}
\usepackage{lipsum}

\newcommand{\skienacut}[1]{}

\title{Gatsby without the `E': Creating Lipograms with LLMs}

\author{
Rohan Balasubramanian \quad Nitish Gokulakrishnan \quad Syeda Jannatus Saba \quad Steven Skiena \\
Stony Brook University \\
\texttt{\{rohan.balasubramanian, nitish.gokulakrishnan, syedajannatus.saba\}@stonybrook.edu} \\
\texttt{skiena@cs.stonybrook.edu}
}

\begin{document}
\maketitle
\begin{abstract}
Lipograms are a unique form of constrained writing where all occurrences of a particular letter are excluded from the text, typified by the novel
%, pose significant challenges in both literary and computational contexts. One famous example is 
\textit{Gadsby} \citep{wright1939gadsby}, which daringly 
avoids all usage of the letter `e'.
In this study, we explore the power of modern large language models (LLMs) by transforming the novel
%F. Scott Fitzgerald's 
\textit{The Great Gatsby}~\citep{fitzgerald1925great} into a fully `e'-less text. 
%The task is particularly challenging given the high frequency of the letter 'e' in the English language, necessitating innovative approaches to maintain fluency and coherence. 
% We employ various generation techniques, starting from simple deletions to more sophisticated methods involving synonym replacement and selective word changes. Our approach emphasizes translation rather than de novo generation to ensure the original text's meaning is preserved.  
We experimented with a range of techniques, from baseline methods like synonym replacement to sophisticated generative models enhanced with beam search and named entity analysis.
% We explore and compare multiple language models for lipogram translation, including T5~\citep{raffel2020exploring}, Llama2~\citep{touvron2023llama}, and Llama3~\citep{llama3modelcard}, evaluating them on metrics such as readability, grammatical correctness, and semantic similarity.
%The results were intriguing: 
%the T5 model stood out for maintaining the original meaning, achieving a cosine similarity score of 0.69 and a low grammatical error rate of 10.39\%. Meanwhile, the Llama3 model dazzled us with its fluency, earning the highest readability score of 102.58.
%\saba{changing it according to the new result}
%
We show that excluding up to 3.6\% of the most common letters (up to the letter `u') had minimal impact on the text’s meaning, although translation fidelity rapidly and predictably decays with stronger lipogram constraints.
Our work highlights the surprising flexibility of English under strict constraints, revealing just how adaptable and creative language can be.

%One  of the most surprising findings was the resilience of the English language.  However, as more letters were omitted, the text quality predictably declined, turning coherent prose into gibberish.
%Our work not only shows that high-quality lipograms can be generated using advanced language models, but it also sets a new benchmark for future research in constrained text generation. 
%\todo{Rewrite the abstract with specific claims from the experiments.}
\end{abstract}

\section{Introduction}

The lipogram is interesting as a highly specific literary form, one that requires the writer to omit all occurrences of a particular letter from the text. Perhaps the best known lipogram is \textit{Gadsby} \citep{wright1939gadsby}, which systematically excludes the letter `e' throughout his novel.
Crafting such a lipogram is challenging, because `e' is the most common letter in English, representing roughly 12\% of the letters appearing in conventional English-language text.

Lipograms provide insight into the flexibility and malleability of language under strong constraints.
Modern large language models (LLMs) excel in generating text under strong stylistic constraints, such as replicating the style of Shakespeare or Dr.\ Seuss.
This makes lipogram generation an excellent stress test for LLMs, to navigate these literary constraints without sacrificing fluency or meaning.

In this paper, we consider the task of producing lipogram ``translations'' of existing texts.
Such translations require clever word substitution and sentence restructuring to satisfy the alphabetic restrictions, all while ensuring that the meaning of the original text is retained in the lipogram version. 

Our particular challenge is to translate F.\ Scott Fitzgerald's classic American novel ``The Great Gatsby'' \cite{fitzgerald1925great} into an `e'-less lipogram.
Gatsby was selected as our target because it is a short (47,094 words) novel familiar to a broad readership, and with the recent expiration of copyright protection now in the public domain.
Finally, the name Gatsby resonates well with {\em Gadsby}, the lipogram novel which is its rough contemporary.

% Table for different approaches
\begin{table*}[t]
\begin{small}
\centering
\begin{tabular}{|>{\centering\arraybackslash}m{0.07\textwidth}|>{\arraybackslash}m{0.85\textwidth}|}
\hline
\textbf{Approach} & \textbf{Generated Text} \\
\hline
Original & In my younger and more vulnerable years my father gave me some advice that I've been turning over in my mind ever since. \\
\hline
E-delete & In my youngr and mor vulnrabl yars my fathr gav m som advic that I'v bn turning ovr in my mind vr sinc. \\
\hline
Synonym & In my jr. and mor vulnrabl days my don afford m about guidance that I'v follow turning ovr in my mind always sinc. \\
\hline
GPT-4o & During my youth, at a time of fragility, my dad imparted wisdom that I continuously mull over in thought. \\
\hline
T5 & I now, in my young, and my most old and as my My dad, though, had a point or two that I'd had in my mind From that point on, I'd had a whirl in my mind.\\
\hline
Llama2 & In my youth and my most fragility , I got from my dad a bit of wisdom which I'm still thinking about and I'm trying to turn it around .\\
\hline
Llama3 & In my days of youth and fragility, it was my dad, who, as I look back on it, has always had a lot of wisdom to pass on to his son. \\
\hline
\end{tabular}

\end{small}
\caption{Representative examples of lipogram translations produced from different generative approaches. \vspace*{-5mm}}
\label{tab:approaches}
\end{table*}

Our contributions in this paper include:

\begin{itemize}[nosep]

    \item \textbf{Methods for Lipogram Translation}:  We present a systematic study of algorithms for creating lipograms, from baseline methods such as synonym replacement to sophisticated generative models enhanced using (i) beam search, (ii) name entity analysis, (iii) training on paraphrased texts, (iv) optimization by sequence trimming and (v) multi-selection.  We present an ablation study that assesses the relative effectiveness of these techniques.

%    \item \textbf{Creating an E-less \textit{The Great Gatsby}}: We successfully transformed \textit{The Great Gatsby} into an e-less version. This involved rigorous evaluation to ensure the paraphrased text stayed true to the original's essence.
   
    \item \textbf{Lipogram Evaluation}: We measure the quality of generated lipogram translations across a variety of metrics, including readability, grammatical correctness, and semantic similarity to the original source text.
    We use this evaluation suite to compare the performance of three different LLMs, % (T5, Llama2, and Llama3), 
    demonstrating that that Llama3 produced the most faithful as well as the most readable constrained translations.
    %These metrics and the data resources we will provide open up a new challenge for future researchers to produce better lipogram translations.
   
    \item \textbf{The Malleability of English Under Strong Constraints}:  We generalized our lipogram translation algorithms to eliminate any specified subset of alphabet symbols from the text.   This provides a laboratory to study the effect of restriction strength (the fraction of letter frequency excluded) on the semantic similarity of translation (measured by cosine similarity). 
    We demonstrate that the elimination of letters of frequency up to `u' (3.6\%) can be done with minimal loss of fidelity, but beyond this range fidelity degrades in a strikingly regular exponential decay, culminating in gibberish once large fractions of the alphabet are excluded.

\end{itemize}

Our project not only demonstrates the feasibility of creating high-quality lipograms using LLMs, but also provides valuable insights into the complexities of constrained text generation.
We provide the first page of our lipogram translation of Gatsby for inspection in Section \ref{sec:lipogram-gatsby} of the appendix, along with the original source text.
\footnote{Upon acceptance of this paper we will release our full lipogram translation and all associated data/code.}

% This paper is organized as follows.
% We survey previous work on lipograms and constrained text generation in Section \ref{sec:previous-work}.
% Our lipogram generation and evaluation methods are presented in Sections \ref{sec:generation-approaches} and \ref{sec:evaluation}, respectively.
% We present our results on the malleability of English in Section \ref{sec:fidelity-analysis}, with suggestions for future work.

\section{Previous Work}
\label{sec:previous-work}

%We review the literary history of lipograms and NLP research on constrained text generation in the following subsections.

%\subsection{Literary and Popular Discussion}

Several literary works have explored the constraints of omitting specific letters, including:

\begin{itemize}[nosep]
    \item \textbf{Gadsby}~\citep{wright1939gadsby} by Ernest Vincent Wright is a lipogram novel that avoids the letter `e'. It tells the story of John Gadsby and his efforts to revitalize the fictional city of Branton Hills. Wright tied the `e' key down on his typewriter to prevent his use of this vowel, although a few instances did slip through.
    
    \item \textbf{La Disparition (A Void)}~\citep{perec1969void} by Georges Perec is a French novel inspired by ``Gadsby" that also excludes the letter `e'. It was later translated into English by Gilbert Adair while maintaining this constraint. 
    % The narrative follows the mysterious disappearance of a certain Anton Vowl, weaving a complex plot without using `e'.
%
     Perec followed up with his novel ``Les Revenentes'' \cite{perec2018revenentes} which uses {\em only} words with the vowel `e', showcasing his exploration of constrained writing beyond lipograms.
    
    % \item \textbf{Eunoia}~\citep{bok2001eunoia} by Christian Bök is a univocalic lipogram, where each chapter uses only one vowel. It explores the creative possibilities of writing with severe constraints, producing distinct and innovative prose.
\end{itemize}

%\subsection{Literary and Popula r Discussion}
The challenge of constrained writing has garnered significant attention in both literary circles and popular media. The Oulipo (Ouvroir de littérature potentielle) group
%, co-founded by Raymond Queneau and François Le Lionnais, 
was instrumental in exploring the potential of literary constraints. Their works
%, including Perec's \textit{La Disparition} and Calvino's \textit{If on a winter's night a traveler}~\citep{calvino1993if}, 
exemplify the creative possibilities that constraints can unlock in literature \cite{calvino1993if,oulipo}.

%\subsection{Other Constraint-Based Human Work}
% Authors have experimented with various other constraints in their writing beyond omiting the letter  `e'. Alphabetical constraints in writing began with Tryphiodorus' adaptation of the Odyssey, written between the 2\textsuperscript{nd} and 4\textsuperscript{th} centuries, omits successive letters of the Greek alphabet in each book, such as `alpha' in the first book, 'beta' in the second, and so on.~\citep{WikipediaLipogramHistory} This is analogous to the lipogram translation task we set out in this paper.

% A related challenge is that of producing palindromic text. There exists a small genre of works such as ``Dr. Awkward \& Olson in Oslo'' \cite{levine1986dr,puder2016dr} where the text reads the same backward as forward.

Recent strides in NLP have redefined constrained text generation. \citet{roush2023most} introduced token filtering for robust constraint adherence, while \citet{kumar2021controlled} framed it as an optimization problem. \citet{zhang2020pointer} proposed POINTER, an insertion-based transformer for efficient hard constraints, and \citet{liu2021constrained} used reinforcement learning to improve concept coverage, though it struggles with stricter constraints. \citet{post2018fast} introduced dynamic beam allocation to enforce lexical constraints efficiently. We build on this by handling stricter character-level bans and ensuring coherent long-form generation.

\section{Generation Approaches}
\label{sec:generation-approaches}

Here we discuss our methods for translation under lipogram constraints.
We will compare them against two simple baselines:

\begin{itemize}[nosep]
 \item \textbf{E-Removal:} Simply removes all instances of `e’ from the text. This approach introduces spelling errors and reduces readability. 
\item \textbf{Synonym Replacement:} Words were lemmatized and replaced with `e’-free synonyms from WordNet~\citep{miller1995wordnet}. If no suitable synonym existed, the original word was retained with `e’s removed. 
\item \textbf{GPT-4o Prompting:} Prompts GPT-4o to rewrite text without the letter `e’; often fluent but may break constraints.
\end{itemize}

Our more sophisticated approaches augment LLMs with several design decisions to improve consistency and coherence over long texts, detailed below.
Table \ref{tab:approaches} presents a representative source paragraph successively translated to lipogram using each of these approaches.

\paragraph{Fine tuning for Paraphrasing:}
To enhance performance on the paraphrasing task, we fine-tuned our model using the PAWS~\citep{zhang2019paws} dataset, which provides pairs of original and paraphrased texts.
For fine-tuning the Llama2 model, we utilized QLoRA~\cite{dettmers2024qlora} with 4-bit quantization and the SFTTrainer framework. 
%The quantization helped in making it possible to run on smaller GPUs without sacrificing performance.
%
For T5, the loss function was modified to include penalties for generating words with the letter `e' and for maintaining high semantic similarity to the original text.
%(Loss\textsubscript{Sim}).
%, as shown in Equation \ref{eq:total_loss}:
%$$
% \begin{equation*}
% \text{Total Loss} = \lambda_1 \times \text{Loss}_\text{E} + \lambda_2 \times \text{Loss}_\text{Sim}
% \label{eq:total_loss}
% \end{equation*}
% where $\lambda_1$ and $\lambda_2$ control the trade-off between avoiding `e' and preserving meaning.

\paragraph{Constrained Beam Search:}
To generate text that strictly avoids the letter `e', we employed a constrained beam search approach during decoding. Tokens containing `e’ were filtered using a \textit{bad\_words\_ids} list, and beam search was configured with 20 beams to expand the search space, and a no-repeat n-gram size of 3 was applied to avoid repetitive patterns in the output.

\paragraph{Multiselection:}
Our model generated multiple candidate translations
% (3 for T5/Llama2, 10 for Llama3),
% \todo{Why different parameters?}
selecting the one with the highest cosine similarity to the original. This ensured compliance with constraints while preserving content and context, reducing randomness or irrelevance.

% Our model produced several different candidate translations (3 for T5 and Llama2, 10 for Llama3) and then we selected the one with the highest cosine similarity to the original text.  This step ensured that the generated text was not only compliant with the constraint but also closely aligned with the original in terms of content and context, minimizing the risk of producing random or irrelevant text.

%Instead of generating a single paragraph translation, 
%To generate text that strictly avoids the letter `e', we employed a constrained beam search approach during decoding. Beam search allows the exploration of multiple potential sequences simultaneously, increasing the likelihood of an optimal output.

%We explicitly filtered out tokens containing the letter `e' by including them in the bad\_words\_ids list, ensuring that no token with the letter `e' could be part of the generated sequence. Key parameters were set to guide the beam search process: the minimum and maximum lengths of the generated sequences were defined to control the output size, and the temperature set to 0.90 to manage the randomness of the sampling process. We specified the number of beams to consider during decoding, set to 20 to widen the search space. Additionally, we configured a no-repeat n-gram size of 3 to avoid repetitive patterns in the output.

\paragraph{LLM Selection:}
We experimented with both encoder-decoder (\textit{T5-small}) and decoder-only (\textit{Llama2-7b-chat-hf} and \textit{Llama3-8b-Instruct}) models. T5 employed constrained beam search, named entity recognition (NER), and pronoun resolution, while Llama2 and Llama3 incorporated QLoRA quantization and constrained decoding. 
%vFor each fine-tuned model, we applied beam search for optimal output selection, and constrained output lengths to 50–150\% of the input to ensure consistency and fidelity.

\paragraph{Post Processing:}
We refined the generated text to ensure clarity and readability. Overlong outputs were truncated to remove irrelevant content. Punctuation was standardized, lengthy lists were trimmed, and empty paragraphs were eliminated to maintain flow. LanguageTool  corrected grammatical errors, while Proselint~\cite{Proselint} polished the writing style and improved spacing.

\begin{table}[t]
\centering
\begin{small}
\begin{tabular}{|l|c|c|c|c|c|c|}
\hline
\textbf{Model} & \textbf{e\%} & \textbf{CS} & \textbf{OOV} &  \makecell{\textbf{GM}} & \textbf{RD} \\
\hline
Original & 37.54 & 1.00 & 35.56 & 1.19 & -3.10 \\
\hline
E-Delete & 0.00 & 0.61 & 51.32 &  29.26 & -2.21 \\
\hline
Synonym & 0.00 & 0.67 & 40.21  & 24.85 & -2.78 \\
\hline
GPT-4o & 5.94 & 0.74 & 40.21  & 1.62 & -2.83 \\
\hline
T5 & 0.00 & 0.69 & 35.27 & 10.39 & -3.15 \\
\hline
Llama2 & 0.00 & 0.65 & 37.41 & 16.40 & -3.75\\
\hline
Llama3 & 0.00 & 0.71 & 36.37 & 1.08 & -3.96\\
\hline
\end{tabular}
\end{small}
\caption{Evaluation metrics on source and lipogram translation text produced from different generative approaches.\vspace*{-3mm}}
\label{tab:model_scores}
\end{table}

\begin{table}[t]
\centering
\begin{small}

\begin{tabular}{|l|c|c|c|c|c|c|}
\hline
\textbf{Model} & \textbf{CS} & \textbf{OOV} & \makecell{\textbf{GM}} & \textbf{RD} \\
\hline
M & 0.69 & 35.04 & 0.88 & -4.00 \\
\hline

M--FT & 0.23 & 63.17 &  26.47 & -2.21 \\
\hline
M--(MS, BS) & 0.42 & 35.74  & 0.93 & -3.62 \\
\hline
% Model--(BeamS, Multiselection) & 0.68 & 28.46 & 0.00  & 5.90 & 86.49 \\
% \hline
M--PS & 0.43 & 36.53 & 2.37 & -4.00 \\
\hline
\end{tabular}
\end{small}

\caption{Ablation studies on our Llama3-based lipogram translation model. M = Complete Model, FT = Finetuning, MS = Multiselection, BS = Beam Search, PS = Post Processing\vspace*{-4mm}}
\label{tab:ablation_studies}
\end{table}

\section{Evaluation Methods}
\label{sec:evaluation}

We evaluated the text on semantic similarity, readability, grammatical accuracy, and constraint adherence. All of our methods achieved 100\% adherence to the `e’-less constraint. Table \ref{tab:model_scores} highlights model performance across these metrics.

% \subsection{Constraint Violations (Presence of `E')}
% We first evaluated adherence to our primary constraint of excluding the letter `e'. This involved 
% %checking the generated text for any instances of the letter `e' and 
% calculating the percentage of violations: 
% \begin{equation*}
% \text{E-Score} = \left( \frac{W_e}{W_t} \right) \times 100
% \end{equation*}
% where
% %\begin{itemize}
% %    \item 
%     \( W_e \) is the number of words having the letter `e' in the generated text,
% %    \item 
% and \( W_t \) is the total number of words in the generated text.
% %\end{itemize}

\begin{figure}[h!]
    \centering
    \includegraphics[width=\linewidth]{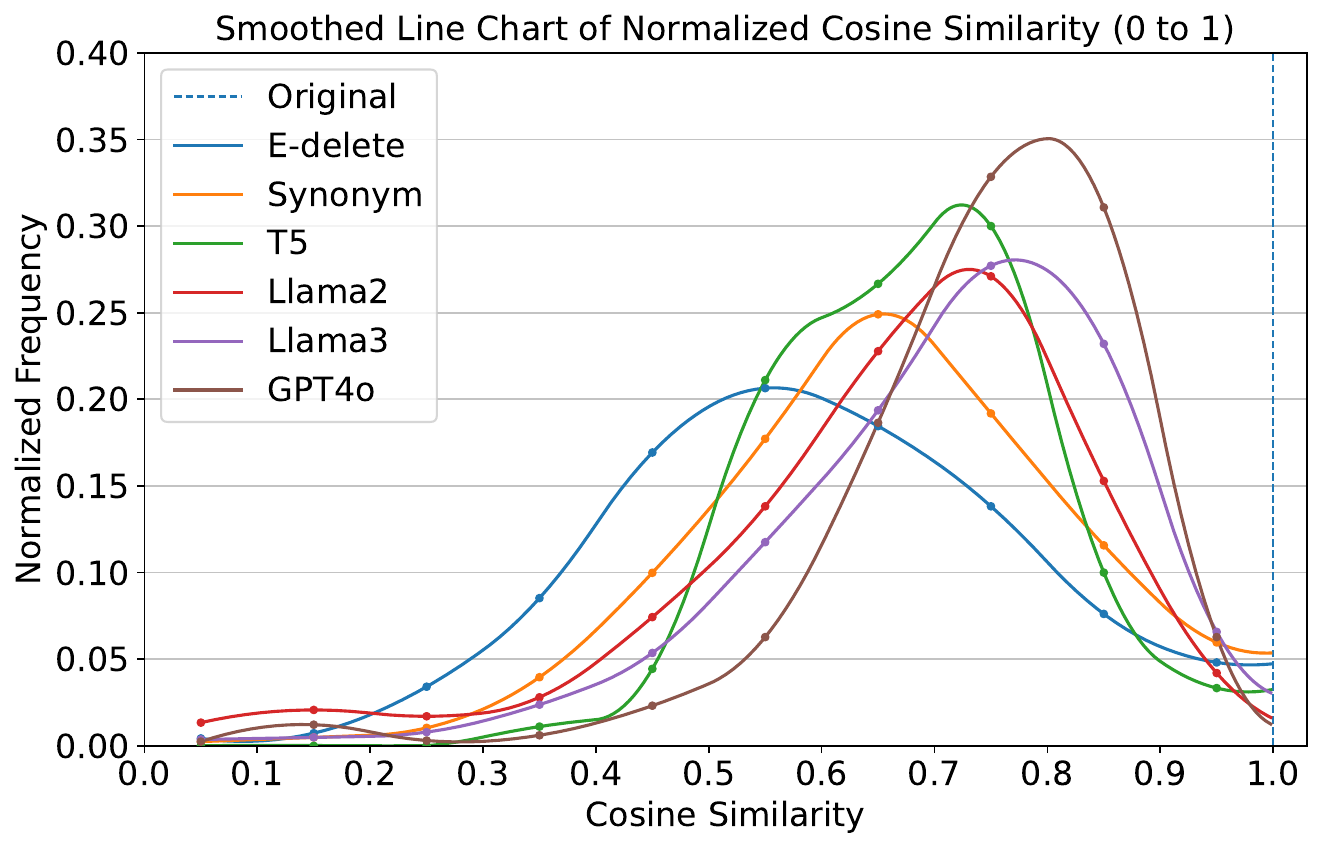} 
    \caption{Cosine similarity distributions for lipogram translations.  Better models have right-shift distributions. \vspace*{-10mm}}
    \label{fig:dist_cosine_similarity}
\end{figure}

\paragraph{Cosine Similarity of Embeddings (CS):}
We used cosine similarity to measure the semantic equivalence between the original source and our generated lipogram translation. 
%This method computes the dot product of embeddings of both texts to quantify how closely they align in terms of semantics.
For embeddings, we utilized the the SentenceTransformer model ``paraphrase-MiniLM-L6-v2" model to encode semantics~\citep{reimers-2019-sentence-bert}. %\saba{add citation for it if you can}
Figure~\ref{fig:dist_cosine_similarity} presents a histogram of paragraph similarity over our models.
% Models producing right-shifted peaks indicate better performance, with
%have spread-out peaks, indicating a broader range of similarity scores. This suggests these models produce outputs with varying degrees of contextual similarity, with 
GPT-4o and Llama3 generate more similar paragraphs, although there remains a sizable gap with the original Gatsby text.

\paragraph{Out of Vocabulary Score:}
Lipogram construction algorithms are naturally tempted to delete all occurrences of forbidden symbols, leaving non-vocabulary word behind (i.e.\ bhind).
To measure this we calculated an out of vocabulary (OOV) score as the percentage of words in the generated text that were not found in standard dictionaries.
% A higher OOV score suggests a greater degree of ill-formed content.
From Table~\ref{tab:model_scores}, we note that 35.56\% of the words in original {\em Gatsby} qualify as OOV, largely reflecting proper nouns and obsolete terms. E-Delete inflates OOV to 51.32\% due to ill-formed words, while T5 (35.27\%) and Llama3 (36.37\%) closely align with the original, reflecting better lexical precision.

% Figure~\ref{fig:dist_oov} plots the per-paragraph OOV frequency for each model.
% The E-delete method performs the worst, with over half the words OOV (and many non-OOV terms matching unrelated words).
% %peaks at 30-40\% OOV, indicating high out-of-vocabulary content in this range. 
% %The 
% T5 best matched the source distribution, suggesting better vocabulary handling, while the Synonym and Llama2 models 
% also perform well.
% Llama3 performs significantly well as visible from the steep rise, but it has a similar distribution to other approaches for higher OOV scores, barring T5. 
% \saba{change the colors of the plots}

% \begin{figure}[h!]
%     \centering
%     \includegraphics[width=\linewidth]{images/oov.pdf} 
%     \caption{OOV score distribution for lipogram models. Better models exhibit left-shifted distributions.}
%     \label{fig:dist_oov}
% \end{figure}

\paragraph{Readability Scores (RD):}
% \begin{figure}[h!]
%     \centering
%     \includegraphics[width=\linewidth]{images/readabilty_score.pdf} 
%     \caption{Readability score per-paragraph distributions.  Left-shifted distributions are declared more readable.}
%     \label{fig:dist_reading_ease}
% \end{figure}
To evaluate the readability of the generated text, we have utilized the TRank Readability Score \cite{trokhymovych2024open}.
%This new multilingual model for Automatic Readability Assessment (ARA) adapts a ranking-based architecture to score individual texts, where a larger score indicates more difficult readability.
Using this metric, we observed that readability is worst for the E-Delete method, followed by the synonym replacement method, with the best readability observed for the Llama3 method, as expected.

% The distributions presented in Figure \ref{fig:dist_reading_ease} shows that original text has an average readabilty score of 82.42, denoting moderately readable text.
% % \todo{Need new number here.}
% The E-delete model peaks in terms of the easiest-to-read content, revealing the limitations of a score based solely on word/sentence lengths and not vocabulary content.
% The Llama2 and Llama3 models also produce readable text,
% %with peaks around 80-90, 
% while Synonym and T5 show a wider range of readability. Overall, the Llama3 model performs best in maintaining text simplicity
% and readability. 
%\saba{put the results with filtered out data on the table}

% \todo{Move the grammatical mistake chart to appendix.}
\paragraph{Grammatical Mistake (GM):}
We assessed the grammatical accuracy of the generated text using Python's LanguageTool \cite{mozgovoy2011dependency}. 
% This tool helped identify and correct grammatical errors, ensuring that the generated text adheres to standard language conventions.
The score is given in terms of the average number of mistakes found in translating each source paragraph, so longer paragraphs are fairly expected to have more errors.
% The paragraph-score distributions are presented in Appendix Figure \ref{fig:dist_grammar_mistake}.
As presented in Table~\ref{tab:model_scores} (clearer in Appendix~\ref{sec:appendix} Figure~\ref{fig:dist_grammar_mistake}), Llama3 closely matches the original text in terms of grammatical correctness, followed by T5 and Llama2.
The baseline E-Delete method introduces the most grammatical errors.
%Llama2 generates text with more grammatical mistakes than T5 and Llama3. 

\subsection{Ablation Study}
\label{sec:ablation}
To better understand how each part of our lipogram generation pipeline contributes to the final output, we systematically removed components and observed the effects on the first 200 paragraphs. Table~\ref{tab:ablation_studies} presents the results. 
% Stripping away our prompt engineering efforts dropped the similarity score to 0.42, but improved readability significantly to 95.75\%. The simpler prompt generates readable sentences but failed to maintain coherency with the paragraph given. This suggests that our finely tuned prompts might need some rethinking to balance readability and coherence.
%
Finetuning on paraphrase data proved important to downstream performance. 
%Without it, the similarity score dropped dramatically to 0.29, and the out-of-vocabulary (OOV) percentage jumped to 48.91\%. This shows just how crucial 
Finetuning keeps the generated text semantically close to the original and reducing odd, out-of-place (OOV) words. Moreover, readability difficulty increased to -2.21, showing the importance of finetuning for maintaining both the semantic integrity and overall quality of the text. 
% \todo{saba: discuss these w/ prof and rewrite accordingly}
Removing multiselection and beamsearch also led the reduction in similarity to 0.42.
%Interestingly, omitting both beam search and multiselection led to the lowest OOV percentage at 28.46\% but increased grammatical mistakes to 5.90\%, revealing a trade-off between vocabulary accuracy and grammatical correctness. 
Excluding post-processing resulted in a marginal increase in the similarity score by 0.01 but caused a higher OOV percentage of 36.53\% and more grammatical mistakes (2.37\%). 

\begin{figure}[h]
    \centering
    \includegraphics[width=0.47\textwidth]{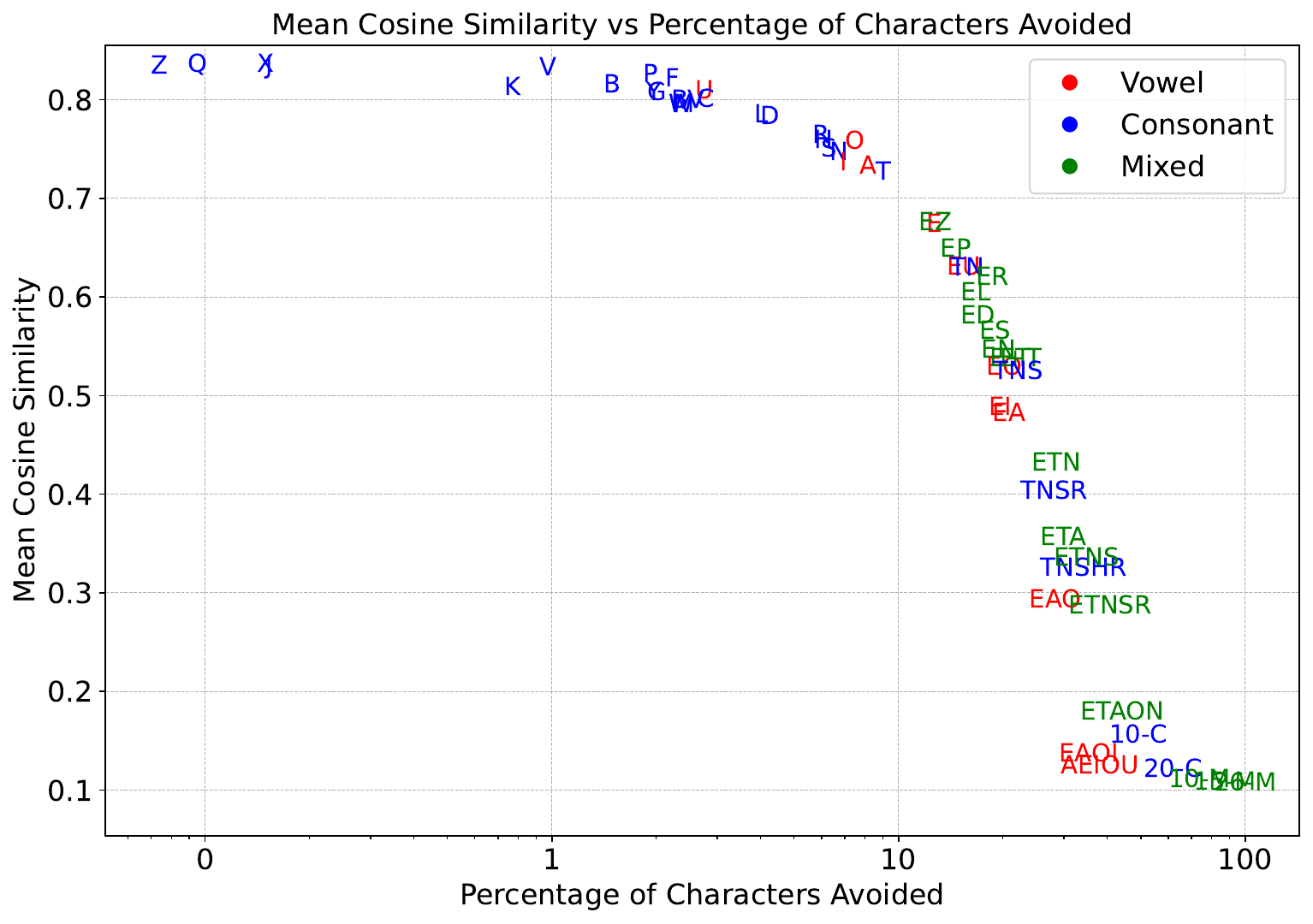} 
    \caption{Relationship between Mean Cosine Similarity and Percentage of Characters Avoided (Log Scale) in Generated Texts.\vspace*{-7mm}}
    \label{fig:gen_const_cosine_log}
\end{figure}
% \vspace{-3mm}

% \todo{I would like to see this figure bigger, perhaps taller if not wider.}

\section{Lipogram Constraints vs Fidelity}
\label{sec:fidelity-analysis}

To explore the malleability of written English, we employed our algorithm to create lipogram versions of {\em Gatsby} under constraints both stronger and weaker than forbidding the letter `e'. We generated lipogram translations for the first 200 paragraphs by excluding each single letter of the alphabet, plus selected subsets of vowels, consonants, and mixed letters to create even greater challenges.
%Representative paragraphs generated for each excluded letter appear in Table \ref{table:avoided_characters}.

Figure \ref{fig:gen_const_cosine_log} presents our results, with the x-axis representing the difficulty of lipogram generation (measured by the exclusion fraction of characters in standard English usage) and the y-axis representing the quality of generation (measured by mean paragraph cosine similarity with the original text).
Each lipogram is represented as a point in restriction-quality space, labeled by the excluded letters.
%
%%Figure ~\ref{fig:cosine_similarity_dist} 
%Figure~\ref{fig:gen_const_cosine_log} presents this data on a log scale, demonstrating that
Weak lipogram constraints are easily overcome.
Eliminating any of the 16 least frequent letters (up to letter `u', with 3.6\% frequency) yield very similar source similarity levels.
Stronger constraints begin to interfere with the story Fitzgerald wanted to tell.

The results show a clear decay trend (clearer in Appendix~\ref{sec:appendix} Figure \ref{fig:cosine_similarity_dist}): the cosine similarity decreases linearly with the exclusion fraction, indicating a steady loss of semantic fidelity even when removing up to 40\% of the character space.
Beyond this point similarity drops to noise levels.
%, albeit greater than zero because our algorithm employs multi-selection.
Excluding single vowels appears no more difficult than consonants of similar frequency, although excluding larger subsets of vowels becomes more crippling.
Every English word must have a vowel: exclude all vowels and you only have numbers and punctuation left to express yourself with. 

%Excluding vowels results in the most significant drops in similarity, highlighting their essential role in maintaining text coherence. While excluding consonants also reduces similarity, the impact is less drastic compared to vowels. Mixed exclusions show intermediate effects, but still trend towards lower similarity as more characters are excluded.

%Overall, the plot underscores that stricter lipogrammatic constraints lead to greater divergence from the original text, with the most severe declines in similarity occurring when a higher percentage of essential characters, particularly vowels are avoided. We can notice as the percentage of characters avoided increases, the graph approaches an asymptote, showing that further exclusions lead to minimal additional loss of similarity. This suggests there is a limit to how much the text can diverge from the original while still maintaining some recognizable structure.

% \begin{figure}[h!]
%     \centering
%     \includegraphics[width=\linewidth]{images/gen_const_grammar_truncated.png} 
%     \caption{Relationship between grammatical mistakes and lipogram constraint in generated texts.}
%     \label{fig:gen_const_grammar}
% \end{figure}

% Sentences may remain semantically similar even as readability degrades. Figure~\ref{fig:gen_const_grammar} plots the corresponding increase in grammatical mistakes per paragraph in lipogram text as word selection gets progressively constrained. Grammatical fidelity largely remains intact up to the exclusion of `u', at less than two mistakes per paragraph before rising sharply with greater restriction.

\section{Conclusion}
%Creating an e-less version of "The Great Gatsby" using advanced language models has been both challenging and insightful. Our project 
Lipogram translation demonstrates the power of modern language models, 
%while highlighting the trade-offs between 
maintaining strict constraints while preserving text quality. 
In exploring several LLMs 
%such as T5, Llama2, and Llama3, 
we identified their respective strengths/weaknesses in handling constrained text generation. This study underscores the complexities of creating high-quality lipograms and provides a benchmark for future research in constrained text generation.
%and lipogram translation.

%Our findings reveal that excluding the letter `e', the most frequent in English, significantly impacts readability and grammatical accuracy. Simple deletion methods often fail to maintain coherence, while more sophisticated techniques like synonym replacement show promise but still struggle with the constraint.

%Left open for future work is the use of LLMs to produce coherent translations under even stronger palindromic constraints, where the text must read the same forward and backwards.
%Norvig~\citep{Norvig-2016} used computer search techniques to produce a 21,012 word palindrome, but with no underlying story or meaning.
%Perhaps techniques akin to those developed in our work can be used produce interesting palindromic translations.
% }

\clearpage

\newpage

\section*{Limitations}
Although our methods produce reasonably satisfying lipogram translations, they have limitations inherent to the complexity of the task.  No reader is likely to prefer our lipogram over the classic original text, except for curiosity sake.
We do not consider lipogram generation in languages other than English, although we believe similar results should be obtainable with similar methods.

%One issue is that sometimes the models arbitrarily remove the letter "e" from words, leading to strange and hard-to-read outputs.  
There remain several problems with our lipogram translations, including maintaining gender consistency, because a character's pronouns or gender references can change within the same context, creating confusion and inconsistency. 
Finally, the models struggle to maintain logical flow and coherence across paragraphs, resulting in a somewhat disjointed and fragmented narrative. Successfully addressing these issues will result in generated text that is more coherent, consistent, and readable.

% Entries for the entire Anthology, followed by custom entries
% \bibliography{anthology,custom}
% \bibliographystyle{acl_natbib}

\newpage
\onecolumn
\appendix

\section{Appendix}
\label{sec:appendix}

\begin{figure}[h!]
    \centering
    \includegraphics[width=\linewidth]{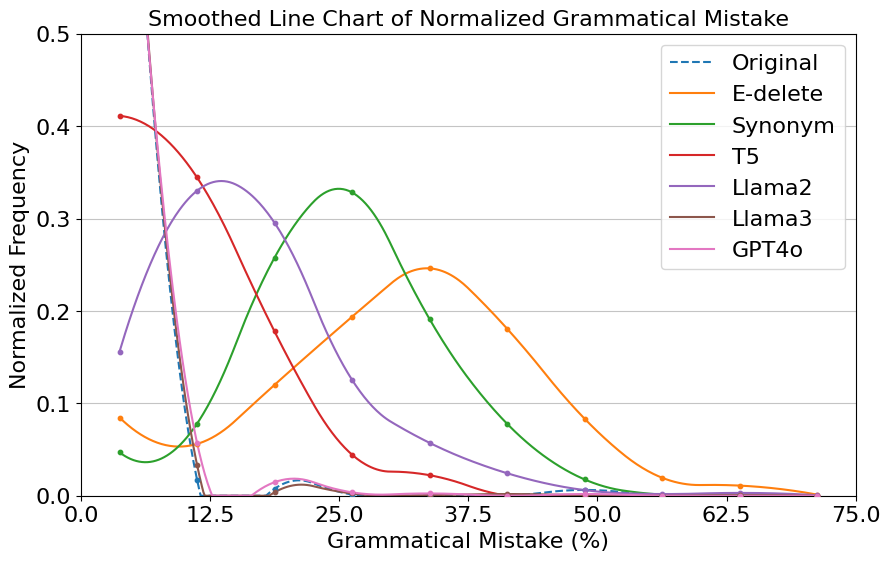} 
    \caption{Grammar mistake distribution for lipogram models. Better models have left-shifted distributions.}
    \label{fig:dist_grammar_mistake}
\end{figure}
\begin{figure*}[h!]
    \centering
    \includegraphics[width=\textwidth]{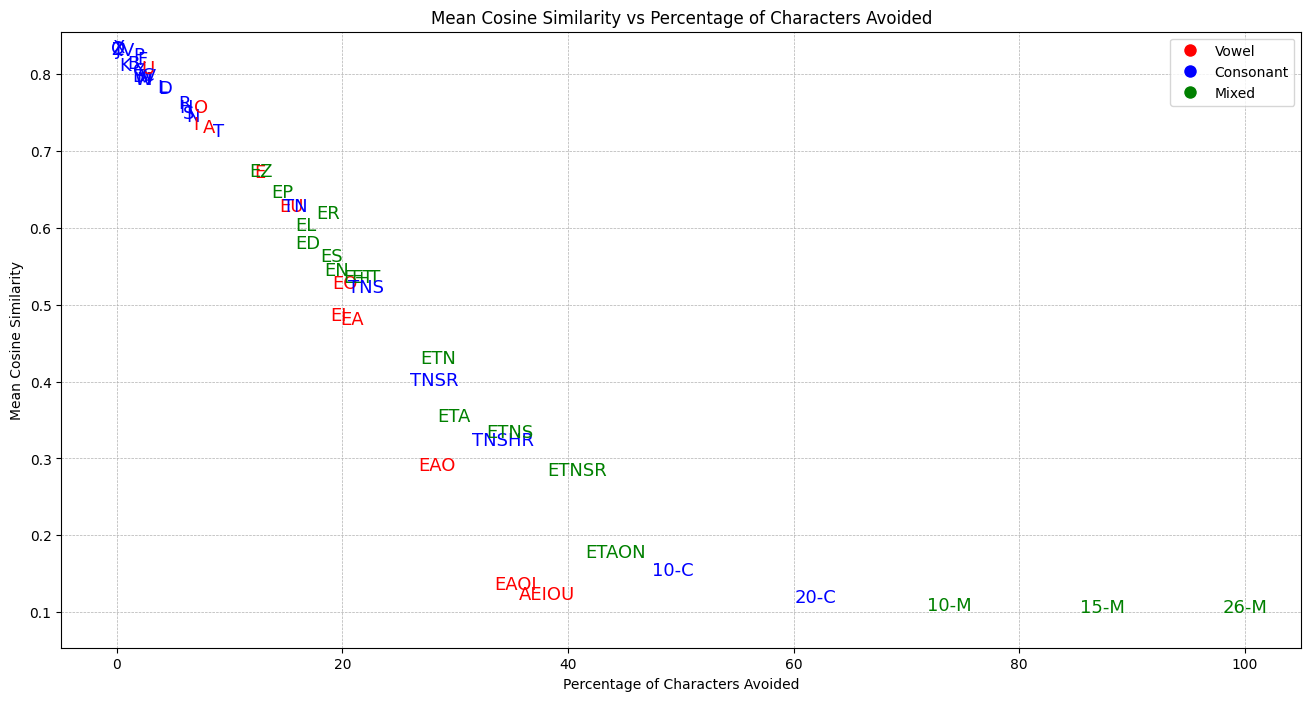} 
    \caption{Relationship between mean cosine similarity and percentage of characters avoided (linear scale) in generated texts under different constraints, illustrating a consistent linear relationship between quality and strength of constraint.}
    \label{fig:cosine_similarity_dist}
\end{figure*}

\subsection*{Examples of Generated Text}

Table \ref{table:avoided_characters} presents examples of the same paragraph from the beginning of {\em The Great Gatsby} translated as lipograms excluding each letter of the alphabet.
Section \ref{sec:lipogram-gatsby} presents our e-less lipogram translation of the first page of {\em Gatsby}, with the corresponding source text available for inspection in Section \ref{sec:original-gatsby}

\begin{longtable}{|>{\centering\arraybackslash}m{0.2\textwidth}|>{\arraybackslash}m{0.7\textwidth}|}
\caption{Generalized Constraint Generation} \label{table:avoided_characters} \\
\hline
\textbf{Characters Avoided} & \textbf{Text Generated} \\
\hline
\endfirsthead

\hline
\textbf{Characters Avoided} & \textbf{Text Generated} \\
\hline
\endhead

\hline \multicolumn{2}{|r|}{\textit{Continued on next page}} \\ \hline
\endfoot

\hline
\endlastfoot

Original & ``Whenever you feel like criticizing anyone", he told me, "just remember that all the people in this world haven't had the advantages that you've had." \\
\hline
A & ``Whenever you think of criticizing someone'', he tells me, ``just remember, everyone in the world didn't get the benefits you got.'" \\
\hline
B & ``Whenever you want to criticize anyone,'' he said to me, `` just keep in mind that everyone in the world hasn't had as many advantages as you've had.", \\
\hline
C & ``Whenever you want to blame anyone'', he said to me, `` just remember all people on this earth have not had advantages you have had." \\
\hline
D & ``Whenever you want to criticize anyone'', he tells me, ``just remember, all people on this earth haven 't been given the same opportunities as you have." \\
\hline
E & ``As soon as you think about criticizing anybody'', I told him, ``just think about how much luck you had that you had not had as much luck as you had.''   \\
\hline
F & ``Whenever you criticize anyone'', he said to me, `` just remember all people on this earth haven't had advantages you've had.'' \\
\hline
G & ``Whenever you want to criticize anyone'', he said to me, ``just remember everyone in the world hasn't had as many opportunities as you've had.'' \\
\hline
H & ``Every time you want to criticize someone'', said to me, ``just remember, all people on our planet did not enjoy as many advantages as you did.'' \\
\hline
I & ``Whenever you want to blame anyone'', he tells me, ``just remember all people on earth have not had advantages you have had.'' \\
\hline
J & ``Whenever you criticize anyone'', he said to me, `` Remember that everyone in the world hasn't had as many advantages as you've had.'' \\
\hline
K & ``Whenever you criticize anyone'', he said to me, `` just remember all people on this earth haven't had advantages you've had.'' \\
\hline
L & "Whenever you want to criticize anyone", he said to me, "remember that everyone on this earth hasn't had as many advantages as you've had. \\
\hline
M & ``Whenever you want to criticize anyone'', he said, ``just think about the fact that everyone in the world hasn't had advantages like you've had. '' \\
\hline
N & ``Remember,'' he said to me, ``that every time you criticize somebody, you have to remember all those people who have the same privileges as you.'' \\
\hline
O & ``Just remember, whenever he tells me, he said : ``All the benefits that I've had haven't had any advantages.'' \\
\hline
P & ``Whenever you criticize anyone'', he said to me, ``Remember that everyone in the world has not had advantages as you have had.'' \\
\hline
Q & ``Whenever you criticize anyone'', he said to me, ``Remember that everyone in the world has not had advantages as you have had.'' \\
\hline
R & ``Anytime you want to blame someone'', he said to me, ``just keep in mind that not all people on this planet have had advantages like you.'' \\
\hline
S & "Whenever you want to criticize anyone'', he warned me, ``Remember that everyone in the world did not have the advantage you had.'' \\
\hline
T & ``Whenever someone feels like judging anyone else'', he said, ``remember, all people have had more chances in life.'' \\
\hline
U & ``Whenever you feel like criticizing anyone", he told me, "just remember that all the people in this world haven't had the advantages that you've had." \\
\hline
V & ``Anytime you want to criticize someone'', he said to me, ``just remember, all people on this earth did not enjoy the benefits you had.'' \\
\hline
W & ``Anytime you criticize someone'', he said to me, ``remember that everyone on this planet hasn't had as many advantages as you've had.'' \\
\hline
X & ``Whenever you want to criticize anyone", he said to me, "remember that everyone in the world has not had advantages as you have had." \\
\hline
Y & ``Whenever someone feels like criticising someone else'', he said to me, ``just remember, all people on this earth haven`t had advantages like the ones I`ve had.'' \\
\hline
Z & ``Whenever you want to criticise anyone',' he said to me, ``just remember all people on this earth have not had advantages you have had.'' \\
\hline
AEIOU & "`` `````` `` ''`` '' '' ``'' '' '``'``' ``''' ''''' ``'''' '''''' ''' ''''''' '''''' ''' ''' `` '''" \\
\hline

\end{longtable}
\newpage
\section{Generated E Less Gatsby}
\label{sec:lipogram-gatsby}

In my days of youth and fragility, it was my dad, who, as I look back on it, has always had a lot of wisdom to pass on to his son.

``As soon as you think about criticizing anybody'', I told him, ``just think about how much luck you had that you had not had as much luck as you had.''

I did not say anything about it, but I always had an unusual way of communicating with him in an uncommunicating way, and that is why I think it is important to maintain all of my opinions. This habit has brought to light a lot of curious minds, which I also had to put up with as a victim, not to say that I wasn't a boor in my own right. Abnormal minds quickly pick up on this trait and cling to it in normal individuals. It was in this way that, during my school days, I got unjustifiably caught up in politics, as I had a knack for picking up wild, unknown man's sorrowful thoughts, most of which had nothing to do with politics. My confidant was usually a plagiarist, with obvious omissions. Holding on to your opinions is an act of faith. I'm still afraid I might miss out on a thing or two, if it wasn't for my dad, who was a bit of a snob in his own way. And I, in turn, am just as snobby as him.

And, having said all this about my pliability, I admit that this limit has. Moral conduct may stand on hard rocks or on swampy grounds, but I do not mind what is standing on it, as long as it stands on that limit. Coming back to India last fall I had a craving for a world that was in a uniform morality and was always on a moral standstill; I did not want to go on any rambunctious trips, and I didn't want any glowing insights into human souls. But only Gatbsy, this man, who is known by his book, was not a part of it. If you look at him, you will find nothing in him that is common or ordinary, nothing from which you can draw an unsatisfactory conclusion about his origins; but if you must know my opinion of him I will say this, that his body is a triumph of gymnastics, a work of art, and so is his soul. His soul is as admirably its body, full of unusual tricks and as startling as an acrobat's body. It is this combination of body and soul, of spirituality and physicality, which was so captivating and intoxicating.

My family was a famous, rich family in that city of this Mid-Atlantic for 3rd. Carroway is a kind of clan, and it has a custom that it is born from duchy of Buccluich, but it's actual originator is my grandpa's cousin, who was born in 1851, had a stand-in for civil war and was to start with a company that sold tools in bulk, which my dad is carrying on to this day.

I don't know if I saw him or not, but it is said that I look similar to him, with a particular allusion to that hard-build painting that is hanging in my dad's room. I was born in 1891, just as my dad was born, and in 1920, I took part in that post-war migration that is known as World War I. I had so much fun that I didn't want to go back, so I thought I would go back to school. All my cousins and aunts thought about it as if I was choosing a boarding school, and finally, said, “I think it's right.” My dad thought it was all right, too, and was willing to pay for my tuition, but I didn't go back until 1922.

It was a practical thing to find a room in town, but I was just coming from a country with a lot of lawns, and it was warm, so I had to go out on my own to a commuting city. A young man from work had found a bungalow for us at \$80 a month, but finally his firm told him to go to Washington. I had an old Ford, a dog, and a woman from Finland, who was my cook and my maid, and who had run away.

It took about a day and a half for a man, who was just arriving than I, to stop on my way.

“How can you go to town?” “I'm asking you,” said Nick. 

I had told him, and as I was walking on, I was solitary again. I was guiding, a trailblazing, an original inhabitant. It was a casual gift from him that I was now a part of this community.

And so, with a lot of sun and a big burst of growth of plants, as with things that grow quickly in films, I was familiar with that conviction again, that it was starting to grow again in spring.

I bought so many books that it was hard to know what to do with it all. It was not only a lot of books, but also many good things to pull down from young air, which was full of vitality. In addition to all this, I also bought 12 books on banks and loans and stocks and bonds, all of which stood in my library in a row, in gold and bright colors, as if it would unfold all that was only known by Morgan, Midus and Macassas. Now I would go back to my old ways and bring all sorts of things that I did not know about, so I could do it again.

\section{Corresponding Original Text from {\em The Great Gatsby}}
\label{sec:original-gatsby}

In my younger and more vulnerable years my father gave me some advice that I've been turning over in my mind ever since.

“Whenever you feel like criticizing anyone,” he told me, “just remember that all the people in this world haven’t had the advantages that you’ve had.”

He didn’t say any more, but we’ve always been unusually communicative in a reserved way, and I understood that he meant a great deal more than that. In consequence, I’m inclined to reserve all judgements, a habit that has opened up many curious natures to me and also made me the victim of not a few veteran bores. The abnormal mind is quick to detect and attach itself to this quality when it appears in a normal person, and so it came about that in college I was unjustly accused of being a politician, because I was privy to the secret griefs of wild, unknown men. Most of the confidences were unsought—frequently I have feigned sleep, preoccupation, or a hostile levity when I realized by some unmistakable sign that an intimate revelation was quivering on the horizon; for the intimate revelations of young men, or at least the terms in which they express them, are usually plagiaristic and marred by obvious suppressions. Reserving judgements is a matter of infinite hope. I am still a little afraid of missing something if I forget that, as my father snobbishly suggested, and I snobbishly repeat, a sense of the fundamental decencies is parcelled out unequally at birth.

And, after boasting this way of my tolerance, I come to the admission that it has a limit. Conduct may be founded on the hard rock or the wet marshes, but after a certain point I don’t care what it’s founded on. When I came back from the East last autumn I felt that I wanted the world to be in uniform and at a sort of moral attention forever; I wanted no more riotous excursions with privileged glimpses into the human heart. Only Gatsby, the man who gives his name to this book, was exempt from my reaction—Gatsby, who represented everything for which I have an unaffected scorn. If personality is an unbroken series of successful gestures, then there was something gorgeous about him, some heightened sensitivity to the promises of life, as if he were related to one of those intricate machines that register earthquakes ten thousand miles away. This responsiveness had nothing to do with that flabby impressionability which is dignified under the name of the “creative temperament”—it was an extraordinary gift for hope, a romantic readiness such as I have never found in any other person and which it is not likely I shall ever find again. No—Gatsby turned out all right at the end; it is what preyed on Gatsby, what foul dust floated in the wake of his dreams that temporarily closed out my interest in the abortive sorrows and short-winded elations of men.

 My family have been prominent, well-to-do people in this Middle Western city for three generations. The Carraways are something of a clan, and we have a tradition that we’re descended from the Dukes of Buccleuch, but the actual founder of my line was my grandfather’s brother, who came here in fifty-one, sent a substitute to the Civil War, and started the wholesale hardware business that my father carries on today.

I never saw this great-uncle, but I’m supposed to look like him—with special reference to the rather hard-boiled painting that hangs in father’s office. I graduated from New Haven in 1915, just a quarter of a century after my father, and a little later I participated in that delayed Teutonic migration known as the Great War. I enjoyed the counter-raid so thoroughly that I came back restless. Instead of being the warm centre of the world, the Middle West now seemed like the ragged edge of the universe—so I decided to go East and learn the bond business. Everybody I knew was in the bond business, so I supposed it could support one more single man. All my aunts and uncles talked it over as if they were choosing a prep school for me, and finally said, “Why—ye-es,” with very grave, hesitant faces. Father agreed to finance me for a year, and after various delays I came East, permanently, I thought, in the spring of twenty-two.

The practical thing was to find rooms in the city, but it was a warm season, and I had just left a country of wide lawns and friendly trees, so when a young man at the office suggested that we take a house together in a commuting town, it sounded like a great idea. He found the house, a weather-beaten cardboard bungalow at eighty a month, but at the last minute the firm ordered him to Washington, and I went out to the country alone. I had a dog—at least I had him for a few days until he ran away—and an old Dodge and a Finnish woman, who made my bed and cooked breakfast and muttered Finnish wisdom to herself over the electric stove.

It was lonely for a day or so until one morning some man, more recently arrived than I, stopped me on the road.

“How do you get to West Egg village?” he asked helplessly.

I told him. And as I walked on I was lonely no longer. I was a guide, a pathfinder, an original settler. He had casually conferred on me the freedom of the neighbourhood.

And so with the sunshine and the great bursts of leaves growing on the trees, just as things grow in fast movies, I had that familiar conviction that life was beginning over again with the summer.

There was so much to read, for one thing, and so much fine health to be pulled down out of the young breath-giving air. I bought a dozen volumes on banking and credit and investment securities, and they stood on my shelf in red and gold like new money from the mint, promising to unfold the shining secrets that only Midas and Morgan and Maecenas knew. And I had the high intention of reading many other books besides. I was rather literary in college—one year I wrote a series of very solemn and obvious editorials for the Yale News—and now I was going to bring back all such things into my life and become again that most limited of all specialists, the “well-rounded man.”

\end{document}